\documentclass[runningheads]{llncs}
\usepackage[T1]{fontenc}
\usepackage[USenglish]{babel}
\usepackage[utf8]{inputenc}
\usepackage{xspace}
\usepackage{csquotes}
\usepackage{amsmath,amsfonts}
\usepackage{amssymb}
\usepackage{comment}
\usepackage{geometry}
\usepackage{rotating}
\usepackage{booktabs}
\usepackage{array}
\usepackage[newcommands]{ragged2e}
\newcommand{\PreserveBackslash}[1]{\let\temp=\\#1\let\\=\temp}
\newcolumntype{L}[1]{>{\PreserveBackslash\raggedright\hyphenpenalty=10000}p{#1}}
\newcolumntype{R}[1]{>{\PreserveBackslash\raggedleft}p{#1}}
\newcolumntype{C}[1]{>{\PreserveBackslash\centering}p{#1}}
\newcolumntype{P}[2]{>{\PreserveBackslash\raggedright\hyphenpenalty=10000}#1{#2}}

\usepackage{afterpage}
\usepackage{pdflscape}
\usepackage{makecell}
\usepackage{hyperref}
\hypersetup{
    colorlinks=true,       % false: boxed links; true: colored links
    linkcolor=blue,          % color of internal links
    citecolor=blue,        % color of links to bibliography
    filecolor=blue,      % color of file links
    urlcolor=blue           % color of external links
}
\usepackage{graphicx}

\usepackage[ruled,vlined]{algorithm2e}
\usepackage[square,numbers,sort&compress]{natbib}
\makeatletter
\renewcommand\bibsection%
{
  \section*{\refname
    \@mkboth{\MakeUppercase{\refname}}{\MakeUppercase{\refname}}}
}
\renewcommand\@biblabel[1]{#1.}
\makeatother

\usepackage[dvipsnames]{xcolor}
\usepackage[normalem]{ulem}
\newcommand{\SuggestEdit}[3][red]{\textcolor{#1}{\sout{#2}}\textcolor{#1}{#3}}
\let\svthefootnote\thefootnote
\newcommand\colorfootnote[2][black]{\def\thefootnote{\color{#1}\svthefootnote}%
  \footnote{\color{#1}#2}\def\thefootnote{\color{black}\svthefootnote}}
\newcommand\RevComment[3][red]{\protect\colorfootnote[#1]{{\textbf{[#2: #3]}}}}

\newcommand\newrevisor[2]{%%
  \colorlet{#1}{#2}
  \expandafter\newcommand\csname#1\endcsname[2]{\SuggestEdit[#1]{##1}{##2}}%%
  \uppercase{\expandafter\newcommand\csname#1}\endcsname[1]{\RevComment[#1]{\MakeUppercase{#1}}{##1}}%%
}
\newcommand\hiderevisor[1]{%%
  \expandafter\renewcommand\csname#1\endcsname[2]{}%%
  \uppercase{\expandafter\renewcommand\csname#1}\endcsname[1]{}%
}

\newrevisor{manuel}{Purple}
\newrevisor{richard}{Blue}
\newrevisor{youngmin}{OliveGreen}

\newcommand{\EGO}{EGO-\hspace{0pt}LS-\hspace{0pt}SVM\xspace}
\newcommand{\ESE}{SEOFUR\xspace}
\newcommand{\SafeMDP}{\textsc{SafeMDP}\xspace}
\newcommand{\SafeOpt}{\textsc{SafeOpt}\xspace}
\newcommand{\SafeOptMC}{\textsc{Safe\-Opt}-\hspace{0pt}\textsc{MC}\xspace}
\newcommand{\psosafeopt}{Swarm-\hspace{0pt}based \textsc{SafeOpt}\xspace}
\newcommand{\StageOpt}{\textsc{StageOpt}\xspace}
\newcommand{\EFI}{\emph{EFI GPC sign}\xspace}
\newcommand{\goose}{GoOSE\xspace}
\newcommand{\SafeExpOptMDP}{\textsc{Safe\-Exp\-Opt}-\hspace{0pt}MDP\xspace}
\newcommand{\failure}{\ensuremath{n^\text{failure}}}

\renewcommand{\vec}[1]{\ensuremath{\mathbf{#1}}}

\let\svthefootnote\thefootnote
\newcommand\blankfootnote[1]{%
  \let\thefootnote\relax\footnotetext{#1}%
  \let\thefootnote\svthefootnote%
}

\begin{document}
\title{Safe Learning and Optimization Techniques: Towards a Survey of the State of the Art}
\blankfootnote{\smaller Please cite as: Kim Y., Allmendinger R., López-Ibáñez M. (2021) Safe Learning and Optimization Techniques: Towards a Survey of the State of the Art. In: Heintz F., Milano M., O'Sullivan B. (eds) Trustworthy AI - Integrating Learning, Optimization and Reasoning. TAILOR 2020. Lecture Notes in Computer Science, vol 12641. Springer, Cham. The final authenticated publication is available online at \url{https://doi.org/10.1007/978-3-030-73959-1_12}.}
\author{Youngmin Kim\inst{1}
  \orcidID{0000-0002-7699-6532}, Richard Allmendinger\inst{1}
  \orcidID{0000-0003-1236-3143}, and \mbox{Manuel López-Ibáñez}\inst{1}\inst{2}
  \orcidID{0000-0001-9974-1295}}
\authorrunning{Y. Kim, R. Allmendinger and M. López-Ibáñez}
\institute{Alliance Manchester Business School, University of Manchester,\newline Manchester M15 6PB, UK
  \and School of Computer Science, University of Málaga, 29071 Málaga, Spain
  \\
%\url{youngmin.kim@postgrad.manchester.ac.uk}\\
\url{{youngmin.kim,richard.allmendinger, manuel.lopez-ibanez}@manchester.ac.uk}
}
\titlerunning{A Survey on Safe Learning and Optimization Techniques}
\maketitle  
\begin{abstract}
  Safe learning and optimization deals with learning and optimization problems
  that avoid, as much as possible, the evaluation of non-safe input points,
  which are solutions, policies, or strategies that cause an irrecoverable loss
  (e.g., breakage of a machine or equipment, or life threat). Although a
  comprehensive survey of safe reinforcement learning algorithms was published
  in 2015, a number of new algorithms have been proposed thereafter, and
  related works in active learning and in optimization were
  not considered.
  This paper reviews those algorithms from a number of domains including
  reinforcement learning, Gaussian process regression and classification,
  evolutionary computing, and active learning. We provide the fundamental
  concepts on which the reviewed algorithms are based and a characterization of
  the individual algorithms. We conclude by explaining how the algorithms are
  connected and suggestions for future research.

 % Safe learning and optimization, as considered in this paper, is loosely about learning a function space efficiently, and discovering a final optimal and feasible input point, respectively, to a problem such that non-safe input points encountered during the search process can be evaluated a limited number of times only. 
 % In recent years, a number of safe learning and optimization algorithms have been proposed. However, to the best of our knowledge, there is no work providing a comprehensive overview of the field. The objective of this paper is to fill this gap by providing a review of safe learning and optimization algorithms designed to tackle black-box expensive problems as present, for example, in closed-loop or hardware-in-the-loop scenarios. Algorithms reviewed originate from areas such as reinforcement learning, active learning, evolutionary computing and Bayesian optimization. We provide the fundamental concepts on which the reviewed algorithms are based, a characterization of the individual algorithms, and finally a discussion around how the algorithms are connected and suggestions for future research paths. 

\end{abstract}

\section{Introduction} \label{sec:intro}

Standard learning and optimization algorithms are generally concerned with
trading off exploration and exploitation of an objective function such that
approximation of the objective function is performed efficiently for learning
problems, and such that an optimal solution, policy and/or strategy is
discovered within as few evaluations as possible for optimization
problems. This paper is concerned with learning/optimization
  problems where the evaluation of \emph{unsafe} solutions imply some significant loss, such as damage of experimental
  equipment or personal injury.

  Safe learning and optimization scenarios typically arise in black-box problems, particularly expensive ones. In \emph{black-box} problems, no explicit mathematical model of the safety constraints is available and the value of the safety constraint function can only be known after a solution has been evaluated. If the number of total evaluations is very limited, from ten to a few thousands, due to limited time or available resources, then the problem is also called \emph{expensive}. Such scenarios arise often when the evaluation of a solution requires a simulation~\cite{FerLopAlb2019asoc} or a real-world experiment, as in closed-loop optimization~\cite{Kno2009closed,Allmendinger2012phd}. This paper provides an overview of the
  research carried out in the area of safe learning and optimization,
  primarily, for black-box and expensive problems.
Algorithms designed for solving expensive (black-box) problems exploit information obtained through a series of expensive evaluations to  select the next input point (i.e., solution) for evaluation. In such problems, an evaluation is akin to executing a physical, chemical or biological experiment, and thus, involves the use of resources, such as raw materials, machines, operators, etc. Problems that require  time-consuming computer simulations can be seen as another example of expensive problems~\cite{SmaMcCAll2011efficient,Allmendinger2012phd,SacDuvMai2018ego-ls-svm,BacHelPic2020gaussian,FerLopAlb2019asoc}. As mentioned above, in expensive problems, the evaluation of an unsafe solution can lead to a waste of resources, such as damage/loss of equipment, of which we may have limited availability.

A large body of research has been carried out around algorithm design for expensive problems.
% Approaches include variations of sequential Design of Experiments (also referred to as Bayesian optimization, surrogate-assisted optimization, meta-models and response surface methods)~\mbox{\cite{ForKea2009surrogate,ShaSweWanAdaFre2016}} and de-randomized heuristics~\mbox{\cite{ShiBac2009niching}}, which reduce a certain degree of randomness in the solution generation process to minimize the risk of wasting expensive evaluations.
%Arguably, Bayesian optimization~\cite{SnoLarAda2012nips}
Arguably, Bayesian optimization (also called surrogate-assisted optimization, meta-model and response surface methods)~\cite{ForKea2009surrogate,SnoLarAda2012nips}
has become the default approach for tackling expensive optimization problems. However, research around non-standard problem features and their implications, such as safety, fairness, and dynamic problem aspects, remains dispersed across the different areas of machine learning, AI and optimization.%\MANUEL{This paragraph needs to change to explain that evaluations use resources and unsafe solutions (that violate constraints) waste resources} 

\citet{GarFer2015saferl} provided a survey on safe reinforcement learning (safe RL), including \textit{constrained criterion}-based optimization, which is relevant to this survey. In \textit{constrained criterion}-based optimization, the safe domain of policies (i.e., of input points) is approximated by several constraints. 
There are several approaches in \textit{constrained criterion}-based optimization. The most typical approach constraints a safety function above a certain threshold. Other approaches evaluate only input points that preserve \emph{ergodicity}~\cite{GarFer2015saferl,MolTeo2012safe} or ensure that evaluations are only allowed when the expected variance of their output does not exceed a certain threshold~\cite{GarFer2015saferl}. Classical \emph{ergodicity} means that we can get from every state to every other state with positive probability. However, in the context of safe RL, the \emph{ergodicity} assumption differentiates between safe and unsafe states, and the concepts of reachability (an input point is safe only if it
can be reached without evaluating any unsafe input points) and returnability (an
input point is safe if there is a sequence of transitions via safe input points that reaches any of a given safe state set) are limited to being able to move between safe states only~\cite{TurBerKra2016safemdp,WacSuiYueOno2018aaai,TurBerKra2019safe}; hence, in the remainder of this paper, \emph{ergodicity} refers to this notion from safe RL.
% \textbf{Youngmin's opinion: Classically, in the context of \textit{constrained criterion}-based optimization, preserving \emph{ergodicity} means that we only evaluate states which ensure an agent to get from every state to every other state~\cite{MolTeo2012safe}. Here, the safety of evaluations is guaranteed by limiting the selection of states to those that preserve \emph{ergodicity} with some properly managed probability ~\cite{MolTeo2012safe}. However, in the context of safe learning that has been developed after the publication of the survey of \citet{GarFer2015saferl}, preserving \emph{ergodicity} is ensured by selecting states that deterministically ensure reachability and returnability (which will be explained later in the paper)~\cite{TurBerKra2016safemdp,WacSuiYueOno2018aaai,TurBerKra2019safe}; hence we will refer to this concept as \emph{ergodicity}.}
%reaching to any state from other states is guaranteed when a certain sequence of input points are evaluated. 

In this paper, we define a safe learning and optimization problem as one subject to one or more safety constraints and/or the preservation of ergodicity as defined above. An input point that violates a safety constraint or ergodicity is deemed \emph{unsafe}. The goal when tackling such problems may be to identify an optimal input point, learn some unknown function, explore a search space or determine the boundaries of the safe input space, while constraining the evaluation of unsafe input points up to a maximum budget, which is often zero if no unsafe evaluations are allowed at all.
  Thus, we review algorithms that aim to satisfy all safety constraints and/or preserve ergodicity, the latter  mostly applies to safe RL algorithms designed for learning tasks. 
Furthermore, we review several related algorithms in active learning and optimization communities such as Bayesian optimization and evolutionary computing, which were not covered in~\cite{GarFer2015saferl}. 

The remainder of the paper is organized as follows. In Section \ref{sec:ps}, we
propose a general formal definition of safe optimization problems that
generalizes many of the formulations found in the literature, and discuss the
scope of this survey. Section \ref{sec:fundamentals} provides a brief summary
of several fundamental concepts related to safe learning and
optimization. %including Markov decision process (MDP), Gaussian process (GP) regression, Evolutionary algorithms (EA), some \textit{L}-Lipschitz continuity-based operators, confidence interval (naive and intersectional) and classifiers.
Section \ref{sec:algorithms} reviews existing safe learning and optimization
algorithms that have been proposed after the survey by
\citet{GarFer2015saferl}, originating from evolutionary computing, active
learning, reinforcement learning and Bayesian optimization. Finally,
Section~\ref{sec:conclusion} discusses links between these algorithms and
provides ideas for future research.

\section{Problem statement}\label{sec:ps}

This section provides a formal definition of a general safe optimization problem and discusses various special cases. 

\newcommand{\DecisionSpace}{\ensuremath{\mathbb{R}^m}}
%\MANUEL{Maybe too late, but if we consider in this section policies and state/action pairs, then the decision space is not \DecisionSpace }

%%%%%%%%%%%%%%%%%%%%%%%%%%%%%
%%% MATHEMATICAL MODEL
%%%%%%%%%%%%%%%%%%%%%%%%%%%%%

In a safe optimization problem, we are given an objective function $f\colon \DecisionSpace \to \mathbb{R}$ that is typically both black-box and
expensive (e.g., costly, time-consuming, resource-intense etc.). The goal is to discover a
feasible and safe $m$-dimensional input point
$\vec{x}=(x_1,\dotsc,x_m) \in \DecisionSpace$ that maximizes the objective
function $f(\vec{x})$ while avoiding the evaluation of \emph{unsafe} input points as much as
possible. The objective function may represent reward, efficiency or cost, and
input points may represent policies, strategies, states/actions (of an
agent/system) or solutions. Formally, 
\begin{equation} \label{eq:gd}
\begin{alignedat}{3}
  \text{Maximize}\quad   & f(\vec{x})\qquad& \vec{x} \in \DecisionSpace\\
  \text{subject to}\quad & g_i(\vec{x}) \leq 0 \qquad& i =1,\dotsc,q\ \\
  & s_j(\vec{x}) \geq h_j  \qquad& j =1,\dotsc,p,\\
% FIXME: for the next version of the model, we should include n^failure in the model
\end{alignedat}
\end{equation}
where $g_i(\vec{x}) \leq 0$ are $q$ \textit{feasibility constraints},
$s_j(\vec{x})$ are $p$ \textit{safety functions}~\cite{Gei2006ecml} and $h_j$ are $p$ \textit{safety thresholds}
(constants). We explicitly separate the safety thresholds from the safety functions since, depending on the application at hand, either the function is black-box~\citep{SuiGotBur2015icml} or the threshold is unknown \emph{a priori}~\citep{SchNguEbe2015sal}, meaning we only know whether a solution is safe during or after its evaluation.
A common special case arises when the safety function is the
same as the objective function, that is, there is only one safety constraint
($p=1$) such that $s_1(\vec{x}) = f(\vec{x}) \geq h$
\cite{SuiGotBur2015icml,TurBerKra2016safemdp,AllKno2011ecta}.

There is a key difference between (hard and soft) feasibility and safety constraints: Feasibility constraints model aspects that are relevant for an input point solution to be of practical use, such as bounds of instrument settings, physical limitations or design requirements. Depending on the application, an input point that violates a feasibility constraint can or cannot be evaluated (hard vs soft constraints)~\cite{LikKoc2007predictive} but if a feasibility constraint is violated by a solution, then this has no serious consequences. On the other hand, evaluating an input point that violates a safety constraint leads to an unsafe situation, e.g., an experimental kit breaks or a human dies. That is, prior to evaluating a solution, an optimizer needs to have as much certainty about its safety status as possible, based either on a continuous measure~\cite{KajIkeHaj2009cec} or binary property of solutions.
%\footnote{An alternative problem definition, considered e.g., in~\cite{KajIkeHaj2009cec}, would model safety (or \emph{risk}) as a continuous measure rather than as a binary property of solutions.} 
It is important to note that an input point can be feasible but unsafe, or vice versa. 

An input point that violates any safety constraint is \emph{unsafe}. The
evaluation of an unsafe input point is counted as a \emph{failure}.  Let us
assume that a given budget of failures is allowed, $\failure \in
\mathbb{N}_0$. After this budget is consumed, the algorithm has failed and
cannot continue. We can distinguish two special cases in the literature. In one
case, evaluating \emph{unsafe} input points encountered during the search
process is endurable only a limited number of
times~\cite{AllKno2011ecta,BacHelPic2020gaussian,SacDuvMai2018ego-ls-svm,KajIkeHaj2009cec,SchNguEbe2015sal,SchHarSka2017safe,SchOrtHart2018safe};
thus, $\failure > 0$. When a \emph{failure} represents a relatively innocuous
event, such as the crash of an expensive
simulation~\cite{BacHelPic2020gaussian,SacDuvMai2018ego-ls-svm}, a relatively
large number of failures may be allowed, but each of them has a cost.
However, in other cases, the assumption is that no \emph{unsafe} input point
should be evaluated ever ($\failure=0$)
\cite{TurBerKra2016safemdp,WacSuiYueOno2018aaai,BiyMarAli2019acc,BerKraSch2016bayesian,BerSchKra2016safe,SuiZhuBur2018stageopt,SuiGotBur2015icml,DuiBerCar2017constrained}.

In some safe optimization and learning problems, a safety constraint can be violated if the associated safety function $s_j(\vec{x})$ cannot be evaluated because the unsafe input point represents an incomplete experiment or expensive simulation~\citep{SacDuvMai2018ego-ls-svm,BacHelPic2020gaussian} or physical damage to the input point~\cite{AllKno2011ecta} that prevents measuring the value of the safety function.
Thus, it is only known whether the constraint was violated, but not by how much. Within the above definition of safe optimization problem (Eq.~\ref{eq:gd}), such cases are equivalent to the following safety constraint:
\begin{equation}
  \label{eq:unknown_violation}
  s'_j(\vec{x}) \geq 0\qquad\text{where}\qquad s'_j(\vec{x}) = \begin{cases}
    s_j(\vec{x}) & \text{if $s_j(\vec{x}) \geq h_j$ (safe)}\\ 
-1 & \text{if $s_j(\vec{x}) < h_j$  (unsafe)}
\end{cases}
\end{equation}
where only the value $s'_j(\vec{x})$ is observable, whereas $s_j(\vec{x})$ is
not directly observable. Moreover, in some contexts~\citep{BacHelPic2020gaussian}, the
evaluation of an unsafe input point may also prevent the objective function (or
feasibility constraints) to be evaluated, even if different from the safety
function.

%%%%%%%%%%%%%%%%%%%%%%%%%%%%%
%%% APPLICABILITY OF THE MODEL
%%%%%%%%%%%%%%%%%%%%%%%%%%%%%
The above optimization model also covers optimal parameter control, if the
problem may be modelled as online optimization or multi-arm
bandit~\citep{BerSchKra2016safe}.
Furthermore, the above model can be adapted easily to safe learning and RL. In
safe learning, the goal becomes to discover a feasible and safe input point
that minimizes the largest amount of uncertainty. Thus, the objective function
may represent uncertainty about a performance or safety function. Uncertainty
may be measured in different ways, e.g., variance and width of confidence
interval.  In the particular case of safe RL, a Markov decision process (MDP)
is typically used for modelling the problem, such that an agent (e.g., rover or
quadrotor) needs to explore the state or state-action space in an uncertain
environment~\cite{TurBerKra2016safemdp,WacSuiYueOno2018aaai,BiyMarAli2019acc,TurBerKra2019safe}.
The sequence of input points (states or state-action pairs) is determined by a
transition function, which is often unknown. In this context, a safe input
point must also satisfy \textit{ergodicity}, i.e., it needs to satisfy
the properties of reachability  and returnability to a safe state~\citep{TurBerKra2016safemdp,TurBerKra2019safe,WacSuiYueOno2018aaai}. Here, returnability is
allowed to be met in $n$ steps, however, reachability is problem-specific, as
we will explain in more detail later. In the context of MDP, the
goal may be to optimize a reward function~\citep{Gos2009rl,WacSuiYueOno2018aaai} or to find the
largest set of safe input points (\emph{safe
  exploration})~\citep{TurBerKra2016safemdp,TurBerKra2019safe,BiyMarAli2019acc}.
In either case, the objective function in Eq.~(\ref{eq:gd}) may be adapted to
reflect those goals.

% In the other case~, an input point is expected to be safe if it can return to a pre-specifed set of safe states in several steps along with an iteratively estimated 'unknown' transition function, which defines results of actions taken at certain states.\MANUEL{I feel that this last sentence is out of place since we discuss returnability in a different paragraph above.: fixed!}

%Lastly, a different type of result caused by safety constraint
%violation may be assumed, such as \emph{risky violation}, where some %violations
%do not necessarily damage an object~\cite{KajIkeHaj2009cec}.\MANUEL{This
%  duplicates the previous footnote}

\section{Fundamental concepts in safe learning and optimization}\label{sec:fundamentals}

Safe learning and optimization is a domain that is spread across several research communities, each approaching the problem from a different angle, though there are overlaps in methodologies. This section provides an overview of the main concepts underpinning the different methodologies including Markov decision process (MDP), Gaussian process regression (GPR), Evolutionary algorithm (EA), \textit{safe set operator} (which examines safety of input points based on $L$-Lipschitz continuity), and \textit{optimistic expander operator} and \textit{optimistic safe set operator} (which is operated based on a safe set established by \textit{safe set operator} followed by naive confidence interval and intersectional confidence interval applied to the operators).
% MANUEL: I commented this because we do not really say introduce them.
%Also, classifiers such as Gaussian process classifier (GPC), LS-SVM and VA~\cite{KajIkeHaj2009cec} are briefly introduced.
In addition, \textbf{classifiers}, such as Gaussian process classifier (GPC)~\cite{BacHelPic2020gaussian,SchNguEbe2015sal} and least-squares support vector machine (LS-SVM)~\cite{SacDuvMai2018ego-ls-svm}, have been used for inferring the safety of input points, however, we do not introduce them here for brevity. 

We distinguish between function exploration in safe learning and safe \textbf{MDP} exploration in RL.
In function exploration, the goal is to learn some unknown function, while
avoiding unsafe input points.
By contrast, in safe MDP exploration, an agent explores the state space of the
MDP. Hence, the problem definition includes a reachability constraint, which ensures that safe input points are reachable in one step of the MDP (one-step
reachability~\cite{TurBerKra2016safemdp,WacSuiYueOno2018aaai}) or in several
steps ($n$-step reachability~\cite{TurBerKra2019safe}), and a returnability constraint, which ensures that the agent can return to any of a given safe state set, usually, in several steps ($n$-step returnability~\cite{TurBerKra2016safemdp,WacSuiYueOno2018aaai,TurBerKra2019safe,BiyMarAli2019acc}).
 Details about RL and
MDPs can be found in~\cite{SutBar2018reinf}.

\textbf{GPR} is a regression method that learns a function using Gaussian processes. GPR can be used for function exploration, exploitation and exploration-\hspace{0pt}exploitation~\cite{Aue2002using,SchSpeKra2018gptut}. Following most recent research in safe learning and optimization, we focus here on learning (function exploration) and optimization (function exploration-\hspace{0pt}exploitation) subject to safety constraints, which we denote as safe learning and safe optimization, respectively. 
Arguably, the majority of safe learning and optimization approaches make use of GPR in one way or the other, primarily to infer (i)~an objective function, as generally used in expensive learning/optimization, and/or (ii)~safety function(s), which provides information about the likelihood of an input point being safe. For general information about GPR, the reader is referred to~\cite{BroCorFre2010tutorial,RasWil2006gp,SchSpeKra2018gptut}.

The literature also reports some applications of \textbf{Evolutionary Algorithms (EAs)}~\cite{Back1996evolutionary} to safe optimization problems. EAs are heuristic optimization methods inspired by biological evolution. Loosely defined, EAs evolve a population of individuals (solutions)  through the application of variation operators (mutation and/or crossover) and selection of the fittest. Naive variation operators involve a great deal of (guided) randomness to generate innovative solutions and thus cover a larger part of the search space. Moreover, naive EAs do not account for the expected mean and uncertainty of a solution before evaluating it, unlike GPR. The combination of being too innovative and unconscious about the safety of a solution prior to evaluation may explain their limited application to problems with safety constraints. 
%A more in depth introduction to EAs can be found, e.g., in~\cite{Back1996evolutionary}.

In the literature on safe learning and optimization, two kinds of methods are mostly used for determining the level of safety of input points: $L$-Lipschitz continuity and classifiers. The concept of \textbf{\textit{L}-Lipschitz continuity} is often used in combination with problems where a known constant safety threshold $h$ is available to the learning/optimization algorithm. An evaluation that yields an output value above the threshold would deem the input point as safe. More formally, given a Lipschitz constant $L$, the \textit{L}-Lipschitz continuity assumption is met if
\begin{equation} \label{eq:lipbasic}
    d(f(\vec{x}),f(\vec{x^\prime})) \leq L \cdot d(\vec{x},\vec{x^\prime}),
\end{equation}
where both $\vec{x}$ and $\vec {x^\prime} \in \DecisionSpace$, and $d(\cdot , \cdot)$ denotes the distance between two input points, typically, the Euclidean distance~\cite{BiyMarAli2019acc}. Then, given a set of safe input points $S \subset \DecisionSpace$, an input point $\vec{x} \in \DecisionSpace$ is deemed safe if $\exists \vec{x}' \in S$ such that
\begin{equation}\label{eq:lip}
    l(\vec{x}') - L \cdot d(\vec{x}',\vec{x})\geq h,
\end{equation}
where $l(\vec{x}')$ is the lower bound of the confidence interval for $\vec{x}'$, and $L$ is the Lipschitz constant~\cite{SuiGotBur2015icml}. Using the lower bound makes the above inequality more strict when compared to using the mean or upper bound, thus preventing unsafe evaluations resulting from noisy measurements (i.e., an input point $\vec{x}$ that satisfies the above inequality is highly likely to be safe even if a noisy measurement affects its output value). We refer to Eq.~\eqref{eq:lip} as \textbf{safe set operator}.

Having derived a set of safe solutions using the safe set operator, next, what is called an \textbf{optimistic expander operator} can be used to select input points from that set. The selected points could potentially expand the safe set by helping to classify additional input points (at least one) as safe in the next iteration. That is, an input point $\vec{x}$, not known to be safe, may be classified as safe if the following condition is satisfied:
\begin{equation}\label{eq:opex}
    u(\vec{x}^\text{safe}) - L \cdot d(\vec{x}^\text{safe},\vec{x}) -\epsilon \geq h,
\end{equation}
where $\epsilon=0$ and the evaluation of the safe input point $\vec{x}^\text{safe}$ gives a value equal to $u(\vec{x}^\text{safe})$, the upper bound of the confidence interval for the safety function of $\vec{x}^\text{safe}$~\cite{SuiGotBur2015icml} (optimistic attitude). The \textbf{optimistic expander operator} would select input points that are expected to be safe and also increase the size of the safe set if evaluated. Finally, an \textbf{optimistic safe set operator} uses the condition (Eq.~\ref{eq:opex}), but $\vec{x}$ can be any input point in $\DecisionSpace$ and $\epsilon$ is the parameter representing the noise of a safety function, constructs an optimistic safe set consisting of $\vec{x}$.

%Using $L$-Lipschitz continuity (Eq.~\ref{eq:lipbasic}) requires to compute an accurate estimate of the Lipschitz constant $L$, as lower values of $L$  will make an algorithm more risky regarding safety, while higher values will make it more risk-averse.

Some algorithms calculate \textbf{naive confidence interval} and \textbf{intersectional confidence interval}~\cite{TurBerKra2019safe}, respectively defined as:
\begin{equation}\label{eq:cinaive}
    l_{t}=\mu_{t-1}(\vec{x})-\beta_t\sigma_{t-1}(\vec{x}),\
    u_{t}=\mu_{t-1}(\vec{x})+\beta_t\sigma_{t-1}(\vec{x})
\end{equation}
\begin{equation}\label{eq:ciinter}
    l_{t}=\max(l_{t-1},\mu_{t-1}(\vec{x})-\beta_t\sigma_{t-1}(\vec{x})),\
    u_{t}=\min(u_{t-1},\mu_{t-1}(\vec{x})+\beta_t\sigma_{t-1}(\vec{x})),
\end{equation}
where $\beta_t$ is a parameter that determines the width of the confidence
interval and $\sigma_{t-1}$ is the predicted standard deviation at point
$\vec{x}$~\cite{SchSpeKra2018gptut} at iteration $t$ of the algorithm.

%% MANUEL: This is not really an introduction or definition, so let's move it earlier and say that we don't introduce them.
% In the literature, \textbf{classifiers}, such as Gaussian process classifier (GPC)~\cite{BacHelPic2020gaussian,SchNguEbe2015sal}, least-squares support vector machine (LS-SVM)~\cite{SacDuvMai2018ego-ls-svm}, and customized approaches, such as the violation avoidance (VA) method~\cite{KajIkeHaj2009cec} (which will be explained in more detail in Section~\ref{sec:ea}), have been suggested for inferring the safety of input points. 

\section{Algorithms}\label{sec:algorithms}

This section provides an overview of existing safe learning and optimization algorithms, and relationships between them (Tables~\ref{tab:restable} and~\ref{tab:OA}). The tables can be seen as a first attempt of a classification of these algorithms.

Table~\ref{tab:restable} classifies the safe learning and optimization algorithms according to their aim, either optimization, which is subdivided by the type of method (EA and/or GPR), or learning, which is subdivided into RL or active learning. For each algorithm, the column \emph{Method} shows key features of the algorithm (PSO denotes particle swarm optimization). We notice that most published works use either $L$-Lipschitz continuity  or a classifier  to infer the safety of input points. VA~\cite{KajIkeHaj2009cec} and \goose~\cite{TurBerKra2019safe} are methods that augment other algorithms to handle the safety of input points.

Table~\ref{tab:OA} divides the algorithms by the environment they assume: MDP or non-MDP. For each algorithm, the column \emph{Initial safe seed} shows whether it requires at least one starting input point known to be safe. The column \emph{Safety likelihood} presents the form in which safety of unobserved input points is predicted, either using labels derived from a classifier (safe vs unsafe) or safety constraint(s) (safe set vs the others) (\emph{Label}), probability of safety estimated from a classifier (\emph{Probability}), or none (safety is not inferred: \emph{NI}). As shown in the table, classifiers are used to both predict safety labels of input points and calculate probability of safety.  
%The column \emph{Extent of violation} shows whether an algorithm is able to measure the violated extent of safety function(s) when unsafe input points are evaluated. In \SafeOpt~\cite{SuiGotBur2015icml}, modified \SafeOpt~\cite{BerSchKra2016safe}, \SafeOptMC~\cite{BerKraSch2016bayesian} and \StageOpt~\cite{SuiZhuBur2018stageopt}, it can be problem-specific,\MANUEL{Don't these algorithms assume that safety violations never occur? If so, then it doesn't really matter if they can observe the extent of the violation, because they never use that value, right?} for example, it may be able to observe the extent, e.g., how much a user disliked recommended movie or a patient felt discomfort, or not, e.g., the user stops using the movie recommender after dissatisfaction or the patient dies from dangerous treatment, when the applications to movie recommender system and clinical experiments appeared in some papers~\cite{SuiGotBur2015icml,SuiZhuBur2018stageopt} are differently defined. 
%
The column \emph{Number of objectives} gives the number of objective functions
considered by the algorithm. For example, the number of objectives that
\goose~\cite{TurBerKra2019safe} can handle is problem-specific as the algorithm
is augmented onto algorithms that do not have a built-in approach to cope with
safety constraints (similar to VA~\cite{KajIkeHaj2009cec}) and thus is driven
by the number of objectives that the underlying algorithm is dealing with.  The
column \emph{Safety constraints} show how many safety constraint(s) can be handled by the
algorithm, noticing that some algorithms are limited to a single \textit{safety constraint}. MDP problems additionally include various forms of
  \emph{ergodicity} as a safety requirement.

%\begin{sidewaystable}
\afterpage{%
  \begin{landscape}
\begin{table} 
\caption{Characterization of existing safe learning and optimization algorithms (Part 1).}\label{tab:restable}
\scriptsize
\begin{tabular}{L{2cm} l L{1.4cm} l L{1.5cm} p{6cm}}
  \toprule
\textbf{Discipline}                          & \bf Paper                                                             & \bf Year                            & \bf Algorithm     & \bf Method                         & \textbf{Comment}                                                                                      \\\midrule
\bf Optimization\\
EA                                           & \citeauthor{KajIkeHaj2009cec} \cite{KajIkeHaj2009cec}                 & \citeyear{KajIkeHaj2009cec}         & VA                & NNs                                   & VA is a classifier augmented onto other EAs replacing their offspring generation process.\smallskip \\
                                             & \citeauthor{AllKno2011ecta}
\cite{AllKno2011ecta}                        & \citeyear{AllKno2011ecta}                                             & TGA, RBS and PHC                    &                                        & Various EAs were proposed.                                                                                                 \\ \\
GPR                                          & \citeauthor{SuiGotBur2015icml} \cite{SuiGotBur2015icml}               & \citeyear{SuiGotBur2015icml}        & \SafeOpt          & $L$-Lipschitz continuity                 & Inspired several  safe algorithms~\cite{BerSchKra2016safe,BerKraSch2016bayesian,SuiZhuBur2018stageopt,TurBerKra2016safemdp,WacSuiYueOno2018aaai,TurBerKra2019safe,DuiBerCar2017constrained}. \smallskip\\
                                             & \citeauthor{BerSchKra2016safe} \cite{BerSchKra2016safe}             & \citeyear{BerSchKra2016safe}       & Modified \SafeOpt & 
                                                  & Lipschitz constant-free model.\smallskip   
                                \\
                                             & \citeauthor{BerKraSch2016bayesian} \cite{BerKraSch2016bayesian}     & \citeyear{BerKraSch2016bayesian}   & \SafeOptMC        & $L$-Lipschitz continuity                    & Multiple safety constraints are dealt with.\smallskip   
                                \\ 
                                & \citeauthor{DuiBerCar2017constrained} \cite{DuiBerCar2017constrained}     & \citeyear{DuiBerCar2017constrained}   & \psosafeopt        & PSO                    & Lipschitz constant-free model.\smallskip   
                                \\ 
                                                                             & \citeauthor{SchHarSka2017safe} \cite{SchHarSka2017safe}           & \citeyear{SchHarSka2017safe}      & SBO               &                                     & Application of SAL~\cite{SchNguEbe2015sal} to optimization problem.\smallskip                                              \\
                                             & \citeauthor{SuiZhuBur2018stageopt} \cite{SuiZhuBur2018stageopt}       & \citeyear{SuiZhuBur2018stageopt}    & \StageOpt         & $L$-Lipschitz continuity                    & Multiple safety constraints are dealt with. Optimization is performed in two independent processes: learning and optimization stages.    \smallskip \\
                                             & \citeauthor{BacHelPic2020gaussian} \cite{BacHelPic2020gaussian}             & \citeyear{BacHelPic2020gaussian} 
                                             & \EFI                                                                  & GPC                             & A problem-specific GPC was proposed.\\ \\
GPR with EA                                  & \citeauthor{SacDuvMai2018ego-ls-svm}~\cite{SacDuvMai2018ego-ls-svm} & \citeyear{SacDuvMai2018ego-ls-svm} & \EGO        & LS-SVM                          & GPR is used combined with EA. \\
                           \midrule
\bf Learning                                 &                                                                       &                                     &                   &                                    & \\
Reinforcement Learning & \citeauthor{TurBerKra2016safemdp} \cite{TurBerKra2016safemdp}         & \citeyear{TurBerKra2016safemdp}     & \SafeMDP          & MDP, GPR, $L$-Lipschitz continuity              & Application of \SafeOpt to exploration task. \smallskip                                                                                                                                       \\
                                             & \citeauthor{WacSuiYueOno2018aaai}\cite{WacSuiYueOno2018aaai}          & \citeyear{WacSuiYueOno2018aaai}     & \SafeExpOptMDP    & MDPs, GPR, $L$-Lipschitz continuity              & Extension of \SafeMDP made to maximize cumulative reward while safely exploring the state space.    \smallskip \\
                                             & \citeauthor{TurBerKra2019safe} \cite{TurBerKra2019safe}               & \citeyear{TurBerKra2019safe}        & \goose            & GPR, \newline $L$-Lipschitz continuity                  & The algorithm is augmented onto other unsafe exploration algorithms.    \smallskip \\
                                             & \citeauthor{BiyMarAli2019acc}\cite{BiyMarAli2019acc} & \citeyear{BiyMarAli2019acc} & \ESE & MDP, \newline $L$-Lipschitz continuity  & Transition functions are unknown.    \smallskip \\                                                        
Active Learning                              & \citeauthor{SchNguEbe2015sal}  \cite{SchNguEbe2015sal}              & \citeyear{SchNguEbe2015sal}        & SAL               & GPR, GPC                               & A problem-specific GPC was introduced. 
\smallskip  \\
                                             & \citeauthor{SchOrtHart2018safe} \cite{SchOrtHart2018safe}           & \citeyear{SchOrtHart2018safe}      & SAL               & GPR                                  & Application of SAL to high pressure fuel supply system (HPFS).\\
 \bottomrule
\end{tabular}
\end{table}
\end{landscape}}
%\end{sidewaystable}

\afterpage{\begin{landscape}
\begin{table} 
\caption{Characterization of existing safe learning and optimization algorithms (Part 2).}\label{tab:OA}
\scriptsize
\begin{tabular}{L{2cm} l L{2cm} L{2cm}   L{2cm}  p{6.2cm}}
  \toprule
\textbf{Environment}     & \bf Algorithm          & \bf Initial safe seed      & \bf Safety likelihood  &   \bf Number of objectives  &\bf Safety constraints  \\\midrule

Non-MDP            & VA\cite{KajIkeHaj2009cec}  & Not required & Label         & Single/ Multiple & One or multiple \smallskip \\
                        & TGA, RBS, and PHC \cite{AllKno2011ecta}   & Not required   & NI            & Single & One \smallskip\\
                        & \SafeOpt \cite{SuiGotBur2015icml}  &  Required &  Label                         & Single  & One                       \smallskip  \\
                        & Modified \SafeOpt\cite{BerSchKra2016safe} & Required    & Label            & Single  & One                \smallskip\\
                        &\SafeOptMC \cite{BerKraSch2016bayesian}     & Required         & Label                 & Single & One or multiple  \smallskip\\  
                        &\psosafeopt \cite{DuiBerCar2017constrained}     & Required         & Label                    & Single & One or multiple \smallskip\\  
                        &\StageOpt\cite{SuiZhuBur2018stageopt}    & Required & Label    & Single &  One or multiple  \smallskip \\
                        & SAL \cite{SchNguEbe2015sal}    & Required          & Label    & Single &  One  \smallskip  \\
                        & SBO \cite{SchHarSka2017safe}          & Required          & Label       & Single  & One \smallskip  \\
                        & SAL \cite{SchOrtHart2018safe}   & Required          & Label    & Single &  One  \smallskip  \\
                        & \EGO \cite{SacDuvMai2018ego-ls-svm}  & Not required    & Probability/ \newline Label         & Single &  One or multiple \smallskip \\
                        & \EFI \cite{BacHelPic2020gaussian} & Not required & Probability         & Single &  One \smallskip\\
                           \midrule
MDP  & \SafeMDP \cite{TurBerKra2016safemdp}  & Required  & Label          & Single & One \textit{safety constraint}\newline \textit{Ergodicity}: One-step reachability and returnability\smallskip\\
                        &\SafeExpOptMDP \cite{WacSuiYueOno2018aaai}  & Required  & Label                     & Single  & One \textit{safety constraint}\newline \textit{Ergodicity}: One-step reachability and returnability \smallskip\\      
                        &\ESE \cite{BiyMarAli2019acc}  & Required  & Label                & Single  & Multiple \emph{safety constraints} \newline \textit{Ergodicity}: Returnability \smallskip\\        
                        &\goose \cite{TurBerKra2019safe}  & Required  & Label                        & Single/Multiple  & One \textit{safety constraint}\newline \textit{Ergodicity}: $n$-step reachability and returnability \smallskip\\                                
 \bottomrule
\end{tabular}
\end{table}
\end{landscape}}

\subsection{Evolutionary algorithms (EAs)}\label{sec:ea}
Arguably, the evolutionary computation community was one of the first to investigate safety issues as defined in this paper. The work of \citet{KajIkeHaj2009cec} in 2009 introduced the violation avoidance (VA) method, a classification tool that is augmented onto an EA, while the work of \citet{AllKno2011ecta} in 2011 investigated the impact of safety issues on different stochastic optimizers (TGA, RBS, PHC). VA is able to deal with either binary or continuous input variables, while the optimizers considered in~\cite{AllKno2011ecta} assumed binary input variables. 
%to then use the findings to provide guidelines on the selection of algorithm parameter settings.  

The purpose of the Violation Avoidance (VA) method~\cite{KajIkeHaj2009cec} is to avoid \emph{risky} evaluations by replacing the ordinary offspring generation process of EAs. It applies the nearest neighbors method (NNs) using a weighted distance to decide over the safety of an input point prior to evaluating it. VA assumes the safety label of an input point to be same as the label of the nearest previously evaluated input point.\footnote{The notion of risk is considered more formally in SAL~\cite{SchNguEbe2015sal}, discussed in Section 4.3.}
%To sum up, as the safety of each of the input points (i.e., the offspring generated by an EA) is decided by the safety label of each of their nearest neighbor, it inheres randomness; thus, a large number of safety violations are allowed.

\citet{AllKno2011ecta} studied reconfigurable, destructible and unreplaceable experimental platforms in closed-loop optimization using EAs. Three types of EAs were investigated varying in the way offspring are generated and the level of collaboration of the individuals in a population. The EAs optimized a single objective function, while avoiding trials that have an output value less than a pre-defined (but unknown to the optimizer) lethal threshold, i.e. the safety threshold ($h_j$) in Eq.~(\ref{eq:gd}). An optimization run was terminated if a predefined number of unsafe input points was evaluated or a maximum number of function evaluations reached. No mechanism was put in place to determine safety of an input point prior to evaluating it, but any evaluated unsafe input point was banned from entering the population as a way to guide the population away from the unsafe region in the search space.

\subsection{Algorithms related to \SafeOpt}\label{sec:safeoptder}

Several safe learning and optimization algorithms are based on GPR, $L$-Lipschitz continuity and set theory. \SafeOpt~\cite{SuiGotBur2015icml} was the first algorithm of this type, and has inspired others to follow, such as modified \SafeOpt ~\cite{BerSchKra2016safe}, \SafeOptMC~\cite{BerKraSch2016bayesian}, \psosafeopt~\cite{DuiBerCar2017constrained}, \StageOpt~\cite{SuiZhuBur2018stageopt}, \SafeMDP~\cite{TurBerKra2016safemdp}, \SafeExpOptMDP~\cite{WacSuiYueOno2018aaai}, Safe Exploration Optimized For Uncertainty Reduction (\ESE~\cite{BiyMarAli2019acc}) and \goose~\cite{TurBerKra2019safe}.

\SafeOpt~\cite{SuiGotBur2015icml} was proposed in 2015 for safe optimization aiming to avoid evaluating any \emph{unsafe} input points altogether during the search, i.e., the number of allowed failures is zero. Roughly speaking, the algorithm uses a GP to model the objective function, which together with a known safety threshold is taken as the safety constraint, and the algorithm selects an input point that has the maximum width of confidence interval among those belonging to a \emph{maximizers set} or \emph{expanders set}. The algorithm constructs a safe set using the \textit{safe set operator} (Eq.~\ref{eq:lip}), and to construct the \emph{expanders set} it uses the \emph{optimistic expander operator} (Eq.~\ref{eq:opex}). That is, given an input point deemed to be safe and whose evaluation could potentially help to classify additional input points as safe, then (i)~it belongs to the \textit{expanders set} and (ii)~the \textit{maximizers set} will comprise input points belonging to the safe set whose upper confidence bound is greater than the greatest lower confidence bound among those calculated for all input points in the safe set. Intuitively speaking, generating \textit{expanders set} and \textit{maximizers set} corresponds to classifying input points in the safe set into (i)~\textit{safe set expansion}, which is expected to expand the safe set at the next iteration, and (ii)~\textit{safe set exploitation}, which is likely to yield high output value at the current iteration, by evaluating an input point in the set at the current iteration. Generally, whether to expand or exploit is decided at each step of the algorithm.
%\MANUEL{why do we talk about \StageOpt in the middle of explaining \SafeOpt? Should this sentence go below when we explain \StageOpt?}
%If a suggested input point is expected to make at least one unsafe input point as safe when evaluated, it belongs to the expanders set, and maximizers set is constitute of input points belonging to the safe set whose upper confidence bound is greater than the greatest lower confidence bound among those calculated for all input points in safe set.\RICHARD{This sentence need to be split up into two, and it needs to be rewritten to make it clearer.} 
\SafeOpt~\cite{SuiGotBur2015icml} uses intersectional confidence intervals (Eq.~\ref{eq:ciinter}), and the same is done by \SafeMDP~\cite{TurBerKra2016safemdp}, \SafeExpOptMDP~\cite{WacSuiYueOno2018aaai}, \StageOpt~\cite{SuiZhuBur2018stageopt} and \goose~\cite{TurBerKra2019safe}. 

In general, algorithms inspired by \SafeOpt~\cite{SuiGotBur2015icml} share the fundamental structure of \SafeOpt~\cite{SuiGotBur2015icml}: Constructing the safe set first, and then using it to construct the \textit{expanders set} and \textit{maximizers set}. However, there are some distinctive features between the algorithms. In 2016, three algorithms were proposed: Modified \SafeOpt~\citep{BerSchKra2016safe}, \SafeOptMC~\citep{BerKraSch2016bayesian} and \SafeMDP~\citep{TurBerKra2016safemdp}.
Unlike the original \SafeOpt approach~\cite{SuiGotBur2015icml}, the modified \SafeOpt~\cite{BerSchKra2016safe} estimates the safe, \textit{maximizers} and \textit{expanders} sets without the specification of a Lipschitz constant. The safe set consists of input points whose lower confidence bound is greater than a safety threshold. To estimate the \emph{expanders} set, the algorithm constructs a GP based on both previously evaluated input points and an artificial data point (with a noiseless measurement of the upper confidence bound) selected from the safe set (as opposed to using previously evaluated input points only as done in the original \SafeOpt approach).
\SafeOptMC can deal with multiple constraints, and \SafeMDP applies \SafeOpt to RL problems. \psosafeopt~\cite{DuiBerCar2017constrained}, proposed in 2017, applies a variant of PSO to \SafeOpt, where multiple independent swarms (sets of input points) are used to construct the \textit{maximizers set} and \textit{expanders set}. The objective function differs across swarms to reflect the different goals when constructing the \textit{maximizers} and \textit{expanders set}. Initialized with initial safe seeds, it updates the safe set only when the input points (also referred to as particles in PSO), found at each run of the PSO, are sufficiently far away from the safe input points in the safe set. Here, input points whose lower confidence bound is greater than a safety threshold are assumed to be safe. The input points in the safe set are used to decide the initial positions of the particles at each iteration. As in typical PSO, the particles move toward new positions by considering their current positions and velocities. 
In 2018, \citet{SuiZhuBur2018stageopt} and \citet{WacSuiYueOno2018aaai} proposed the algorithms \StageOpt and \SafeExpOptMDP, respectively. \StageOpt can  deal with multiple safety constraints and divides the process in two independent stages: Safe region expansion stage and optimization stage corresponding to \textit{safe set exploration} and \textit{safe set exploitation}~\cite{SuiZhuBur2018stageopt}, respectively. \SafeExpOptMDP is an extension of \SafeMDP (e.g., unsafe set, uncertain set, etc), and its goal is to maximize the cumulative reward rather than \emph{safe set expansion} (i.e., the safety function is not the objective function). Lastly, in 2019, the RL community proposed the algorithm \goose~\cite{TurBerKra2019safe}, which is augmented onto other unsafe algorithms as a way to control the selection of safe input points.

Algorithms able to handle problems with multiple safety constraints, e.g., \SafeOptMC~\cite{BerKraSch2016bayesian} and \StageOpt~\cite{SuiZhuBur2018stageopt}, model the individual safety constraints, as well as the objective function, using independent GPs. Then, the safe set is the intersection of safe sets estimated from the safety functions, where the \textit{safe set operator} (Eq.~\ref{eq:lip}) is applied to each safety function separately. %The set of potential maximizers contains all safe parameters that could obtain the maximum value given the highprobability boun
Now let us remind that the \textit{maximizers set} is used for \textit{safe set exploitation}, meaning this set is constructed based on the objective function values of input points in the current safe set (ignoring information from the safety functions). Since there is one objective function only, the algorithms \SafeOpt and \SafeOptMC construct the \textit{maximizers set} using the same approach. However, it is not required for \StageOpt, as it has its own independent stage devoted for \textit{safe set exploitation}, and optimization is dealt with by input point selection criterion in that stage. Lastly, \textit{optimistic expander operator} (Eq.~\ref{eq:opex}) is applied to each safety constraint, and all of them should be met when including an input point from the safe set into the \textit{expanders set}. However, \SafeOptMC and \StageOpt apply the structure in slightly different ways, e.g., they define Lipschitz constant(s) and safety threshold(s) differently. %\RICHARD{This paragraph would be nice to show in a flowchart of pseudocode, where it is can easliy be seen which steps are common between the algorithms and which ones not (in that case one should see 'if algorithm = XXX then YYY'.}
When \SafeOpt is applied to RL problems~\cite{TurBerKra2016safemdp,WacSuiYueOno2018aaai}, the set of states allowed to visit is restricted by \textit{ergodicity}. This means that the suggested states should be reachable in one-step and returnable to states in the safe set, established at the previous iteration, in several steps. However, \goose~\cite{TurBerKra2019safe} applies $n$-step expansion until convergence of safe set and \emph{expanders set} at each algorithm iteration by applying $n$-step reachability and returnability. \goose is augmented onto an unsafe RL or Bayesian optimization algorithm (i.e., one that is not designed to deal with safety constraints). The unsafe algorithm suggests an input point $\vec{x}^*$ that belongs to an \textit{optimistic safe set}. Then, \goose evaluates input points from the \emph{expanders set}, until $\vec{x}^*$ is inferred to be safe. Otherwise, \goose asks for another input point from the unsafe algorithm. In \goose, however, additional conditions are used with \emph{optimistic expander operator} for constructing \emph{expanders set}.

%, and works as follows: (i) an unsafe algorithm suggest an input point within a certain sub-domain, ii) if the input point does not fall into the safe set \emph{safe set operator}, an input point, selected from a \textit{expanders set}, is evaluated to update a safety function, (iii) until the input point falls into the safe set, iterate updating the safety function. Please note that \emph{expanders set} was constructed in a different way with \SafeOpt (e.g., noise parameter is considered by \emph{optimistic expander operator}, use of priority function, etc).  
%\goose uses a modified version of \textit{optimistic expander}, which subtract a parameter $\epsilon$ from the left side of Eq.~(\ref{eq:opex}) to cope with noisy measurements ($\epsilon$ uncertainty).
%\goose can also be applied to Bayesian optimization.  and uses modified version of \textit{optimistic expander}, which subtract a parameter $\epsilon$ from the left side of Eq.~(\ref{eq:opex}) to cope with noisy measurements ($\epsilon$ uncertainty).

\ESE~\citep{BiyMarAli2019acc} is an algorithm for safe exploration of
deterministic MDPs with unknown transition functions. Starting from a known safe state set and a list of actions that connect the safe states, and assuming that the unknown transition function is $L$-Lipschitz continuous over both states and actions, the algorithm tries to efficiently and safely explore the search space. The learned (known) transitions are represented in form of a (deterministic) transition function, while an uncertain transition function is defined to handle unknown states. The uncertain transition function maps each state-action pair to all of its possible outcomes, and if there is a state-action pair whose possible outcomes constitute a subset of the known safe state set, then the state is deemed safe. This problem-specific \emph{safe set expansion} method is repeated at each algorithm iteration until it converges ($n$-step returnability). The algorithm removes uncertainty as much as possible at each iteration by greedily optimizing an expected uncertainty reduction measure. The paper~\cite{BiyMarAli2019acc} applies \ESE to a simulation problem with safety constraints.

\subsection{Safe learning and optimization with a classifier}\label{sec:SO}

In addition to the VA method~\cite{KajIkeHaj2009cec} (Section~\ref{sec:ea}), there are three more algorithms that use a classification method for safety inference of input points: Safe Active Learning (SAL)~\cite{SchNguEbe2015sal}, \EGO~\cite{SacDuvMai2018ego-ls-svm} and  \EFI~\cite{BacHelPic2020gaussian}. In particular, \EGO~\cite{SacDuvMai2018ego-ls-svm} and \EFI~\cite{BacHelPic2020gaussian} were proposed to avoid simulations from crashing.

SAL~\cite{SchNguEbe2015sal} is an algorithm for learning a regression model
when unknown regions of the input space can be unsafe. SAL builds two GPs to
approximate an objective function and a discriminative function, mapped to the
unit interval to describe the class (i.e., safe or unsafe) likelihood, for a
problem-specific Gaussian Process classifier (GPC).  SAL assumes that each
evaluated input point $\vec{x}$ provides two additional output values: A
negative (unsafe) or positive (safe) label $c(\vec{x}) \in \{-1, +1\}$ and the
value of a black-box, possibly noisy function $h\colon \mathbb{X} \to (-1,1)$,
where $\mathbb{X} \subset \mathbb{R}^m$ for $m$-dimensional input points. The
function $h(\vec{x})$ provides a noisy \emph{risk} value for evaluated safe
points $\vec{x} \in \mathbb{X}$ (i.e., $c(\vec{x})=+1$) close to the unknown boundary of the safe input
region, while it provides no useful information for unsafe points, i.e.,
$c(\vec{x})=-1$.  SAL also assumes an upper bound for the expected number of
failures.
%
% SAL assumes \manuel{that
%   values of additional output variable , related to safety, are observable when
%   input points are evaluated which are encoded to a}{the availability of
%   black-box} possibly noisy function $h\colon\mathbb{X} \to [-1,1]$, where
% $\mathbb{X} \subset \mathbb{R}^m$ for $m$-dimensional input point
% $\vec{x}$. \manuel{Here, -1 and 1 represent the safe and unsafe class label
%   values, respectively.  If the value of $h$ is greater than a certain risky
%   threshold, a safe class label is assigned to a safe input point when it was
%   evaluated. However, if the value of $h$ is smaller than the threshold (risk
%   detection), then the value of $h$ is given, and the boundary of the safe
%   region (also called decision boundary) is approximated by the input points in
%   this risky region.}{The function $h(\vec{x})$ returns $+1$ for known safe
%   points, $-1$ for evaluated unsafe points, and a noisy \emph{risk} value
%   within $(-1,1)$ for evaluated safe points close to the unknown boundary of
%   the safe input region}. Also, an upper bound for the expected number of
% failures is anticipated.
%
At each iteration, the algorithm selects an input
point with the highest conditional variance given previous observations among
those expected to be safe according to the GPC. Safe Bayesian Optimization
(SBO)~\cite{SchHarSka2017safe} is fundamentally the same as SAL with the core
difference being that, at each iteration of the algorithm, it selects an input
point whose lower confidence bound is the minimum, assuming minimization of the
loss function. Another difference between SAL and SBO is that in SBO a standard GP regression, which is trained with discriminative function values only, is used for the discriminative function, while the use
of class labels (safe vs unsafe) for training GPC is also an option for
SAL. Training a standard GP with discriminative function values only was also the approach
adopted in~\cite{SchOrtHart2018safe}, where SAL was applied to safe learning of
a high pressure fuel supply system.

\EGO~\cite{SacDuvMai2018ego-ls-svm} has been designed for safe optimization of complex systems, where simulations are subject to abrupt terminations due to unphysical configurations, ill-posed problems or lack of numerical robustness.  
That is, it assumes that there is a non-computable sub-domain in the search domain that cannot be expressed by inequality constraints, and thus, applies a binary classifier (least-squares support vector machine or LS-SVM) for this sub-domain. However, the classifier may be used with some inequality constraints that define other non-computable sub-domains.
\EGO constructs an independent GP to model an objective function, and assigns observations into the \textit{safe} or \textit{unsafe class}, which represent computable and non-computable input points, respectively.  Based on the safety labels attached to the previous observations, LS-SVM predicts the probability that an input point belongs to the safe/unsafe class. The paper~\cite{SacDuvMai2018ego-ls-svm} also proposed four different selection criteria for the next input point that combine this probability with the augmented expected improvement (AEI) acquisition function~\cite{HuaAllNot2006global}, which is an input point selection function for Bayesian optimization. Another interesting aspect of the algorithm is that it uses the Covariance Matrix Adaptation Evolutionary Strategy (CMA-ES)~\citep{HanOst2001ec} to estimate the hyper-parameters of the kernel function for GP regression and for LS-SVM.

Lastly, \EFI~\cite{BacHelPic2020gaussian} was also designed to prevent
simulation crashes; thus, it aims to progress an optimization process by
efficiently avoiding input points that are likely to fail. In this context, the
safety function is binary (safe/unsafe) and a failure implies that the
objective function cannot be evaluated. % (i.e., are non-computable).
The approach differs from \EGO in that it cannot construct multiple models
for safety constraints. \EFI constructs a GP for learning binary
inputs, representing safe/unsafe evaluations, that is different from the
classical GPC and more appropriate for deterministic safety functions. In
addition, GPR is used to model the objective function by keeping only
the safe input points.
Finally, \EFI selects an input point by
considering the probability of non-failure multiplied by the standard expected
improvement acquisition function.

\section{Discussion and future research}\label{sec:conclusion}

In this paper, we reviewed and contextualized 18 algorithms designed for safe learning and optimization.

Two studies in the area of evolutionary computing~\cite{KajIkeHaj2009cec,AllKno2011ecta} proposed algorithms (VA~\cite{KajIkeHaj2009cec} and modified versions of EAs~\cite{AllKno2011ecta}) designed for safe optimization. In particular, VA~\cite{KajIkeHaj2009cec} is a flexible approach that can be augmented onto other EAs.

Except for \ESE~\cite{BiyMarAli2019acc} and the aforementioned EAs, the other safe learning and optimization algorithms reviewed in this survey are based on GP regression. However, we are able to observe a trend and can divide these algorithms into categories. The first division we observe is that algorithms adopt the $L$-Lipschitz continuity assumption~\cite{SuiGotBur2015icml,BerKraSch2016bayesian,SuiZhuBur2018stageopt,TurBerKra2016safemdp,WacSuiYueOno2018aaai,TurBerKra2019safe,BiyMarAli2019acc}, use the lower confidence bound~\cite{BerSchKra2016safe,DuiBerCar2017constrained}, or apply a classifier to measure the safety of input points~\cite{BacHelPic2020gaussian,SacDuvMai2018ego-ls-svm,SchNguEbe2015sal,SchHarSka2017safe,SchOrtHart2018safe}. Interestingly, it was found that modified \SafeOpt~\cite{BerSchKra2016safe} and \psosafeopt~\cite{DuiBerCar2017constrained} share a similar structure with algorithms using the $L$-Lipschitz continuity assumption, but are free from  deciding the Lipschitz constant. Furthermore, SEOFUR~\cite{BiyMarAli2019acc} is based on a method that applies the $L$-Lipschitz continuity assumption for the problems where transition functions are unknown. Also, while some algorithms~\cite{SchOrtHart2018safe,SuiGotBur2015icml,BerSchKra2016safe,TurBerKra2016safemdp,WacSuiYueOno2018aaai,TurBerKra2019safe,BacHelPic2020gaussian,SchNguEbe2015sal,SchHarSka2017safe} deal with one \textit{safety constraint}, others~\cite{BerKraSch2016bayesian,SuiZhuBur2018stageopt,SacDuvMai2018ego-ls-svm,DuiBerCar2017constrained} can handle multiple ones. Lastly, while \textit{ergodicity} and \textit{safety constraint}(s) were classified as different concepts applied to distinctive approaches in a previous survey in 2015~\cite{GarFer2015saferl}, we observed that they are used together in \SafeMDP~\cite{TurBerKra2016safemdp}, \SafeExpOptMDP~\cite{WacSuiYueOno2018aaai} and \goose~\cite{TurBerKra2019safe}.

%In standard learning and optimization problems, stopping criterion of an algorithm can vary by user who defines it. However, in the context of safe learning and optimization, we introduce some criteria: fixing the number of evaluations~\cite{SchNguEbe2015sal,AllKno2011ecta} and terminating when noise strength of the function is greater than the maximum width of confidence interval of input points in expanders set, and expanders or maximizers set for exploration~\cite{TurBerKra2016safemdp,WacSuiYueOno2018aaai} and optimization~\cite{SuiGotBur2015icml} problems, respectively. Also, the former can be used for stopping criterion of expansion stage, if an algorithm is operated in two independent stages for \textit{safe set expansion} and \textit{safe set exploitation}. Lastly, one may stop an algorithm if the maximum value computed from an acquisition function for input points in domain becomes smaller than a certain threshold, when Bayesian optimization is used~\cite{SacDuvMai2018ego-ls-svm}. 

Given the above observations, we envision several open questions for future research. First, how to estimate the Lipschitz constant and safety threshold when these are \textit{a priori} unknown. Second, are there real-world applications that would benefit/require alternative formulations of the problem, such as safety thresholds being a function of the input variables and/or change over time, i.e., use $h_j(\vec{x})$, $h_j(t)$ or $h_j(\vec{x},t)$ instead of a fixed constant $h_j$. Finally, it is not clear how to apply multiple safety constraints to MDP problems.
%\MANUEL{but then it will be a safety function, not a threshold, no?} 

%the application of linear safety constraints models, when is this setup more appropriate than constant form constraints, e.g., we can imagine a linearly increasing physical limit of a machine with the increase (or decrease) of the values of input variables\RICHARD{Provide a refernece to the paper where this is done; I have forgotten which paper it was.}. 

%\MANUEL{I think one open question is whether exists a general formulation that covers all variants of safe learning and optimization studied in the literature. Another interesting question is which algorithms are applicable to different variants of the problem that they were not designed for. Another question is whether we can create a benchmark set of safe learning and optimization problems and compare the existing algorithms. Also to study when some of their assumptions do not hold.} 
%Finally, how the algorithms made for safe exploration for reinforcement learning assuming given transition functions can be applied to the opposite problems. \ESE~\cite{BiyMarAli2019acc} included action in input variable, as well as state. This, that is assuming unknown transition functions, may fit to many real-problems; thus, it is useful to investigate how to apply the methods used in majority of safe exploration algorithms e.g., safety threshold, and building safe set, expanders set and \textit{maximizers set} based on Lipschitz continuity, to problems where transition functions are not given.

\vspace*{-1em}
\begin{footnotesize}
\subsubsection*{Acknowledgements.}
% \vspace*{-1em}
M.~L\'opez-Ib\'a\~nez is a ``Beatriz Galindo'' Senior Distinguished Researcher (BEAGAL 18/00053) funded by the Spanish Ministry of Science and Innovation.
%(MICINN).
\end{footnotesize}

\bibliographystyle{splncs04abbrvnat}
\renewcommand{\doi}[1]{doi:\hspace{.16667em plus .08333em}\discretionary{}{}{}\href{https://doi.org/#1}{\urlstyle{rm}\nolinkurl{#1}}}
\renewcommand{\doi}[1]{}
%\bibliography{optbib/abbrev,optbib/journals,optbib/authors,optbib/biblio,optbib/crossref} %,Reference.bib}
\providecommand{\MaxMinAntSystem}{{$\cal MAX$--$\cal MIN$} {A}nt {S}ystem}
  \providecommand{\Rpackage}[1]{#1} \providecommand{\SoftwarePackage}[1]{#1}
  \providecommand{\proglang}[1]{#1}

\end{document}